\documentclass{article}

\usepackage{times}
\usepackage{graphicx} \usepackage{subfigure} 

\usepackage{natbib}

\usepackage{algorithm}
\usepackage[noend]{algorithmic}

\usepackage{hyperref}

\usepackage[accepted]{icml2017}

\usepackage{amsmath}
\usepackage{amssymb}
\usepackage{amsthm}
\usepackage{bbold}
\usepackage{color}
\usepackage{tikz}
\usepackage{wrapfig}
\usepackage[footnotesize]{caption}
\usepackage{booktabs}
\usepackage{multicol}
\usepackage{balance}

\newcommand{\reals}{\mathbb{R}}
\newcommand{\expect}{\mathbb{E}}
\newcommand{\indicate}{\mathbb{1}}

\newcommand{\states}{\mathcal{S}}
\newcommand{\actions}{\mathcal{A}}
\newcommand{\transitions}{P}
\newcommand{\options}{\mathcal{B}}
\newcommand{\tasks}{\mathcal{T}}
\newcommand{\task}{\tau}
\newcommand{\data}{\mathcal{D}}

\newcommand{\astop}{\textsc{stop}}

\newcommand{\added}[1]{#1}

\newcommand{\mycomment}[1]{\STATE\textit{// #1}}

\icmltitlerunning{Modular Multitask Reinforcement Learning with Policy Sketches}
\title{Modular Multitask Reinforcement Learning
with Policy Sketches}

\begin{document} 

\twocolumn[
\icmltitle{Modular Multitask Reinforcement Learning with Policy Sketches}

\icmlsetsymbol{equal}{*}

\begin{icmlauthorlist}
\icmlauthor{Jacob Andreas}{cal}
\icmlauthor{Dan Klein}{cal}
\icmlauthor{Sergey Levine}{cal}
\end{icmlauthorlist}

\icmlaffiliation{cal}{University of California, Berkeley}

\icmlcorrespondingauthor{Jacob Andreas}{jda@cs.berkeley.edu}

\vskip 0.3in
]

\printAffiliationsAndNotice{}  
\begin{abstract} 
  We describe a framework for multitask deep reinforcement learning guided by
  \emph{policy sketches}. Sketches annotate tasks with sequences of named
  subtasks, providing information about high-level structural relationships among tasks
  but not how to implement them---specifically
  not providing the detailed guidance used by much previous work on learning
  policy abstractions for RL (e.g.\ intermediate rewards, subtask completion
  signals, or intrinsic motivations).
  To learn from sketches, we present a model that associates every subtask with a 
  modular subpolicy, and jointly maximizes reward over full task-specific policies by 
  tying parameters across shared subpolicies. Optimization is accomplished via a 
  decoupled actor--critic training objective that facilitates learning common
  behaviors from multiple dissimilar reward functions. 
  We evaluate the effectiveness of our approach in three environments
  featuring both discrete and continuous control, and with sparse rewards
  that can be obtained only after completing a number of high-level subgoals.
  Experiments show that using our approach to learn policies
  guided by sketches gives better performance
  than existing techniques for learning task-specific or shared policies,
  while naturally inducing a library of interpretable primitive behaviors that
  can be recombined to rapidly adapt to new tasks.
\end{abstract} 

\section{Introduction}

\begin{figure}
  \includegraphics[width=\columnwidth,trim=0.2in 3in 3.5in 0.3in, clip]{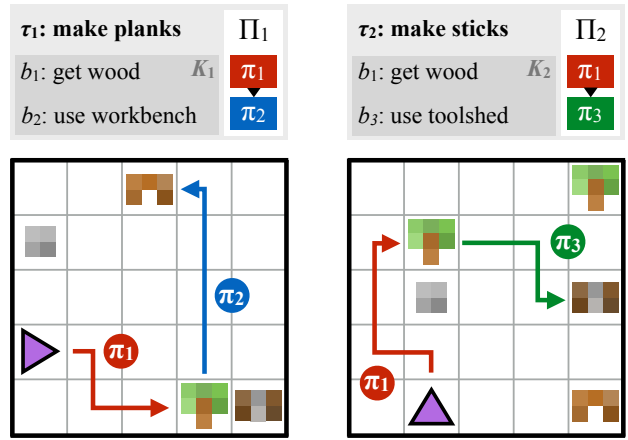}
  \caption{
    Learning from policy sketches. The figure shows simplified versions of
    two tasks (\textit{make planks} and \textit{make sticks}, each
    associated with its own 
    policy ($\Pi_1$ and $\Pi_2$ respectively).
    These policies share an initial high-level action $b_1$: both require the
    agent to \textit{get wood} before taking it to an appropriate crafting
    station. Even without prior information about how the associated behavior $\pi_1$ 
    should be implemented, knowing that the agent should initially follow the same 
    subpolicy in both tasks is enough to learn a reusable representation of their 
    shared structure.
  }
  \label{fig:teaser}
  \vspace{-1em}
\end{figure}

\added{
This paper describes a framework for learning composable deep subpolicies in a
multitask setting, guided only by abstract sketches of high-level behavior.
General reinforcement learning algorithms allow agents to solve 
tasks in
complex environments. But tasks featuring extremely delayed rewards or other
long-term structure are often difficult to solve with flat, monolithic policies,
and a long line of prior work has studied methods for learning hierarchical
policy representations
\citep{Sutton99Options,Dietterich00MaxQ,Konidaris07Skills,Hauser08Primitives}.
While unsupervised discovery of these hierarchies is possible
\citep{Daniel12HREPS,Bacon15OptionCritic}, practical approaches often require
detailed supervision in the form of explicitly specified high-level actions,
subgoals, or behavioral primitives \cite{Precup00Options}.  These
depend on state representations simple or structured enough that
suitable reward signals can be effectively engineered by hand.

But is such fine-grained supervision actually necessary to 
achieve the full benefits of
hierarchy? Specifically, is it necessary to explicitly ground high-level 
actions into the representation of the environment? Or is it sufficient to simply inform the 
learner about the abstract \emph{structure} of policies, without ever
specifying how high-level behaviors should make use of primitive percepts or actions?

To answer these questions, we explore a multitask reinforcement learning setting where the
learner is presented with \emph{policy sketches}.
Policy sketches are short, ungrounded, symbolic representations of a 
task that describe its component parts, as illustrated in \autoref{fig:teaser}. While
symbols might be shared across tasks (\emph{get wood} appears in sketches for 
both the \emph{make planks} and \emph{make sticks} tasks), the learner is told nothing about
what these symbols \emph{mean}, in terms of either observations or intermediate rewards.

We present an agent architecture that learns from policy sketches by
associating each high-level action with a parameterization of a low-level subpolicy, and jointly optimizes
over concatenated task-specific policies by tying parameters across shared subpolicies. 
We find that this architecture can use  the high-level guidance provided by 
sketches, without any grounding or concrete definition, to  dramatically accelerate learning  
of complex multi-stage behaviors. Our experiments indicate that many of the benefits to 
learning that come from highly detailed low-level supervision (e.g.\ from subgoal rewards) can
also be obtained from fairly coarse high-level supervision (i.e.\ from policy sketches). 
Crucially, sketches are much easier to produce: they require no 
modifications to the environment dynamics or reward function, and can be easily provided by
non-experts. This makes it possible to extend the
benefits of hierarchical RL to challenging environments where it may not be possible to specify 
by hand the details of relevant subtasks.
We show that our approach substantially
outperforms purely unsupervised methods that do not provide the learner with any task-specific
guidance about how hierarchies should be deployed, and further that the specific use of sketches to
parameterize modular subpolicies makes better use of sketches than conditioning
on them directly.

}

The present work may be viewed as an extension of recent approaches for
learning compositional deep architectures from structured program descriptors
\citep{Andreas16DNMN,Reed15NPI}. Here we focus on learning in interactive
environments. This extension presents a variety of technical challenges, requiring
analogues of these methods that can be trained from sparse,
non-differentiable reward signals without demonstrations of desired system behavior.

Our contributions are:
\begin{itemize}
  \item A general paradigm for multitask, hierarchical, deep reinforcement
    learning guided by abstract sketches of task-specific policies.     \item A concrete recipe for learning from these sketches, built
      on a general family of modular deep policy representations
      and a multitask actor--critic training objective. \end{itemize}

The modular structure of our approach, which associates every high-level
action symbol with a discrete subpolicy, naturally induces a library of 
interpretable policy fragments that are easily recombined.
This makes it possible to evaluate our approach under a variety of different 
data conditions: (1)  learning the full
collection of tasks jointly via reinforcement, (2) in a zero-shot setting where
a policy sketch is available for a held-out task, and (3) in a adaptation
setting, where sketches are hidden and the agent must learn to adapt a pretrained 
policy to reuse high-level actions in a new task. In all cases, our approach 
substantially outperforms previous approaches based on explicit decomposition of 
the Q function along subtasks \cite{Parr98HAM,Vogel10SARSA}, unsupervised option 
discovery \cite{Bacon15OptionCritic}, and several standard policy gradient 
baselines.

We consider three families of tasks: a \mbox{2-D} Minecraft-inspired crafting
game (\autoref{fig:tasks}a), in which the agent must acquire particular
resources by finding raw ingredients, combining them together in the proper
order, and in some cases building intermediate tools that enable the agent to
alter the environment itself; a 2-D maze navigation task that requires the agent
to collect keys and open doors, and a 3-D locomotion task (\autoref{fig:tasks}b) 
in which a quadrupedal robot must actuate its joints to traverse a narrow winding 
cliff.

In all tasks, the agent receives a reward only after the final goal is
accomplished. For the most challenging tasks, involving sequences of four or
five high-level actions, a task-specific agent initially following a random
policy essentially never discovers the reward signal, so these tasks cannot be
solved without considering their hierarchical structure.
We have released code at \url{http://github.com/jacobandreas/psketch}.

\section{Related Work}

The agent representation we describe in this paper belongs to the broader
family of hierarchical reinforcement learners. As
detailed in \autoref{sec:learning},
our approach may be viewed as an instantiation of the \emph{options} framework
first described by \citet{Sutton99Options}. A large body of work describes
techniques for learning options and related abstract actions, in both single-
and multitask settings. Most
techniques for learning options rely on intermediate supervisory signals, e.g.\ to encourage
exploration \citep{Kearns02Exploration} or completion of pre-defined subtasks
\citep{Kulkarni16DeepHierarchical}. An alternative family of approaches employs
post-hoc analysis of demonstrations or pretrained policies to extract reusable
sub-components \citep{Stolle02LearningOptions, Konidaris11SkillTrees, Niekum15Demonstrations}.
Techniques for learning options with less guidance than the present work include
\citet{Bacon15OptionCritic} and \citet{Vezhnevets16STRAW}, and other general
hierarchical policy learners include \citet{Daniel12HREPS},
\citet{Bakker04Hierarchical} and \citet{Menache02QCut}.  \added{We will see that
  the minimal supervision provided by policy sketches results in (sometimes
  dramatic) improvements over fully unsupervised approaches, while being
substantially less onerous for humans to provide compared to the grounded
supervision (such as explicit subgoals or feature abstraction hierarchies) used
in previous work.}
 
Once a collection of high-level actions exists, agents are faced with the problem
of learning meta-level (typically semi-Markov) policies that invoke appropriate
high-level actions in sequence \citep{Precup00Options}. The learning problem we
describe in this paper is in some sense the direct dual to the problem of
learning these meta-level policies: there, the agent begins with an inventory
of complex primitives and must learn to model their behavior and select among
them; here we begin knowing the names of appropriate high-level actions but
nothing about how they are implemented, and must infer implementations (but not,
initially, abstract plans) from context.  
\added{Our model can be
combined with these approaches to support a ``mixed'' supervision condition
where sketches are available for some tasks but not others (\autoref{ssec:generalization}).}

Another closely related line of work is the Hierarchical Abstract Machines (HAM)
framework introduced by \citet{Parr98HAM}. Like our approach, HAMs begin with a
representation of a high-level policy as an automaton (or a more general
computer program; \citeauthor{Andre01ALISP}, \citeyear{Andre01ALISP}; \citeauthor{Marthi04ALISP}, \citeyear{Marthi04ALISP}) and use reinforcement learning to fill in
low-level details. 
Because these approaches attempt to
learn a single representation of the Q function for all subtasks and contexts,
they require extremely strong formal assumptions about the form of the reward
function and state representation \citep{Andre02ALISPAbstraction} that the
present work avoids by decoupling the policy representation from the value
function. \added{They perform less effectively when applied to arbitrary state representations
where these assumptions do not hold (\autoref{ssec:multitask}). We are additionally
unaware of past work showing that HAM automata can be automatically inferred
for new tasks 
given a pre-trained model, while here we show that it is easy to solve the corresponding
problem for sketch followers (\autoref{ssec:generalization}).}

Our approach is also inspired by a number of recent efforts toward compositional
reasoning and interaction with structured deep models. Such models have been
previously used for tasks involving question answering
\citep{Iyyer14Factoid,Andreas16DNMN} and relational reasoning
\citep{Socher12Semantic}, and more recently for multi-task, multi-robot transfer
problems \citep{Devin16NMN}.  In the present work---as in existing approaches employing
dynamically assembled modular networks---task-specific training signals are
propagated through a collection of composed discrete structures with tied
weights. Here the composed structures specify time-varying
policies rather than feedforward computations, and their parameters must be
learned via interaction rather than direct supervision. Another closely related
family of models includes neural programmers \citep{Neelakantan15NP} and
programmer--interpreters \citep{Reed15NPI}, which generate discrete
computational structures but require supervision in the form of output actions
or full execution traces.

\added{
We view the problem of learning from policy sketches as complementary to the
instruction following problem studied in the natural language processing literature.
Existing work on instruction following focuses on mapping from natural language strings
to symbolic action sequences that are then executed by a hard-coded interpreter 
\citep{Branavan09PG,Chen11Navigation,Artzi13Navigation,Tellex11Commands}.
Here, by contrast, we focus on learning to execute complex actions given symbolic
representations as a starting point. 
Instruction following models may be viewed as joint policies over instructions 
and environment observations (so their behavior is not defined in the absence of
instructions), while the model described in this paper naturally supports adaptation to 
tasks where no sketches are available. We expect that future work might combine the
two lines of research, bootstrapping policy learning directly from
natural language hints rather than the semi-structured
sketches used here.
}

\section{Learning Modular Policies from Sketches}
\label{sec:learning}

We consider a multitask reinforcement learning problem arising from a family of
infinite-horizon discounted Markov decision processes in a shared environment.
This environment is specified by a tuple $(\states, \actions, \transitions,
\gamma)$, with $\states$ a set of states, $\actions$ a set of low-level actions,
$\transitions : \states \times \actions \times \states \to \reals$ a transition
probability distribution, and $\gamma$ a discount factor. Each task $\task \in
\tasks$ is then specified by a pair $(R_\task, \rho_\task)$, with $R_\task : \states \to
\reals$ a task-specific reward function and $\rho_\task : \states \to \reals$ an
initial distribution over states. For a fixed sequence $\{(s_i, a_i)\}$ of
states and actions obtained from a rollout of a given policy, we will denote
the empirical return starting in state $s_i$ as $q_i := \sum_{j=i+1}^\infty
\gamma^{j-i-1} R(s_j)$.  In addition to the components of a standard multitask RL
problem, we assume that tasks are annotated with \emph{sketches} $K_\task$, each
consisting of a sequence $(b_{\task 1}, b_{\task 2}, \ldots)$ of high-level
symbolic labels drawn from a fixed vocabulary $\options$.

\subsection{Model}
\label{ssec:model}

We exploit the structural information provided by sketches by constructing for
each symbol $b$ a corresponding \emph{subpolicy} $\pi_b$. 
By sharing
each subpolicy across all tasks annotated with the corresponding symbol, our
approach naturally learns the shared abstraction for the corresponding subtask,
without requiring any information about the grounding of that task to be
explicitly specified by annotation.

At each timestep, a
subpolicy may select either a low-level action $a \in \actions$ or a
special $\astop$ action. We denote the augmented state space $\actions^+ :=
\actions \cup \{\astop\}$. At a high level, this framework is agnostic to the 
implementation of subpolicies: any function that takes a representation of the 
current state onto a distribution over $\actions^+$ will do.

In this paper, we focus on the case where each $\pi_b$ is represented as a neural network.\footnote{
For ease of presentation, this section assumes that these
subpolicy networks are independently parameterized. As
described in \autoref{ssec:envs}, it is also possible to
share parameters between subpolicies, and introduce discrete
subtask structure by way of an \emph{embedding} of each
symbol $b$.
}
These subpolicies may be viewed as options of the kind described by
\citet{Sutton99Options}, with the key distinction that they have no initiation
semantics, but are instead invokable everywhere, and have no explicit
representation as a function from an initial state to a distribution over final
states (instead implicitly using the $\astop$ action to terminate).

\begin{algorithm}[t]
  \begin{algorithmic}[1]
    \STATE $\data \gets \emptyset$
    \WHILE{$|\data| < D$}
      \mycomment{sample task $\tau$ from curriculum (\autoref{sec:curriculum})}
      \STATE $\task \sim \textrm{curriculum}(\cdot)$ 
      \mycomment{do rollout}
      \STATE $d = \{(s_i, a_i, (b_i=K_{\task,i}), q_i, \task), \ldots\} \sim \Pi_\task$
      \STATE $\data \gets \data \cup d$
    \ENDWHILE
        \mycomment{update parameters}
    \FOR{$b \in \options, \task \in \tasks$}
      \STATE $d = \{(s_i, a_i, b', q_i, \task') \in \data : b' = b, \task' = \task\}$
      \mycomment{update subpolicy}
      \STATE $\theta_b \gets \theta_b + \frac{\alpha}{D} 
        \sum_d \big(\nabla \log \pi_b(a_i|s_i)\big)\big(q_i - c_\task(s_i)\big)$
      \mycomment{update critic}
      \STATE $\eta_\task \gets \eta_\task + \frac{\beta}{D}
        \sum_d \big(\nabla c_\task(s_i)\big)\big(q_i - c_\task(s_i)\big)$
    \ENDFOR
  \end{algorithmic}
  \caption{$\textsc{train-step}(\mathbf{\Pi}, \textrm{curriculum})$}
  \label{alg:inner-loop}
\end{algorithm}

Given a fixed sketch $(b_1, b_2, \dots)$, a task-specific policy $\Pi_\task$ is formed by
concatenating its associated subpolicies in sequence. In particular, the
high-level policy maintains a subpolicy index $i$ (initially $0$), and executes
actions from $\pi_{b_i}$ until the $\astop$ symbol is emitted, at which point
control is passed to $\pi_{b_{i+1}}$.  We may thus think of $\Pi_\task$ as
inducing a Markov chain over the state space $\states \times \options$, with
transitions: \begin{align*}
  (s, b_i) &\to (s', b_i) &\textrm{with pr.}\quad& {\textstyle \sum_{a \in
  \actions}} \pi_{b_i}(a | s) \cdot \transitions(s' | s, a)\\
  &\to (s, b_{i+1}) &\textrm{with pr.}\quad& \pi_{b_i}(\astop | s)
\end{align*}
Note that $\Pi_\task$ is semi-Markov with respect to projection
of the augmented state space $\states \times \options$ onto the underlying state
space $\states$. We denote the complete family of task-specific policies
$\mathbf{\Pi} := \bigcup_\tau \{\Pi_\task\}$, and let each $\pi_b$ be an arbitrary
function of the current environment state parameterized by some weight vector
$\theta_b$. The learning problem is to optimize over all $\theta_b$ to maximize
expected discounted reward \[J(\mathbf{\Pi}) := \sum_\task
J(\Pi_\task) := \sum_\task \expect_{s_i \sim \Pi_\task} \big[ \sum_i \gamma^i
R_\task(s_i) \big]\] across all tasks $\tau \in \tasks$.

\begin{algorithm}[t]
  \caption{\textsc{train-loop}()}
  \label{alg:outer-loop}
  \begin{algorithmic}[1]
    \mycomment{initialize subpolicies randomly}
    \STATE $\mathbf{\Pi} = \textsc{init}()$
    \STATE $\ell_\textrm{max} \gets 1$
    \LOOP
      \STATE $r_\textrm{min} \gets -\infty$
      \mycomment{initialize $\ell_\textrm{max}$-step curriculum uniformly}
      \STATE $\tasks' = \{ \task \in \tasks : |K_\task| \leq \ell_\textrm{max} \}$
      \STATE $\textrm{curriculum}(\cdot) = \textrm{Unif}(\tasks')$
      \WHILE{$r_\textrm{min} < r_\textrm{good}$}
        \mycomment{update parameters (\autoref{alg:inner-loop})}
        \STATE $\textsc{train-step}(\mathbf{\Pi}, \textrm{curriculum})$
                \STATE $\displaystyle \textrm{curriculum}(\task) \propto \indicate[\task \in
        \tasks'] (1
        - \hat{\expect} r_\task) \quad \forall \task \in \tasks$
        \STATE $r_\textrm{min} \gets \min_{\task \in \tasks'} \hat\expect r_\tau$
      \ENDWHILE
      \STATE $\ell_\textrm{max} \gets \ell_\textrm{max} + 1$
    \ENDLOOP
  \end{algorithmic}
\end{algorithm}

\begin{figure}[b]
  \centering
  \includegraphics[width=0.75\columnwidth, trim=0.1in 5in 4.8in 0.2in, clip]{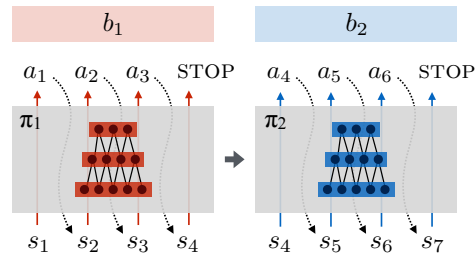}
  \caption{
    Model overview. Each subpolicy $\pi$ is uniquely associated with a symbol
    $b$ implemented as a neural network that maps from a state $s_i$ to
    distributions over $\actions^+$, and chooses an action $a_i$ by sampling
    from this distribution. Whenever the $\astop$ action is sampled, control
    advances to the next subpolicy in the sketch.
  }
  \label{fig:model}
\end{figure}

\subsection{Policy Optimization}

Here that optimization is accomplished via a simple decoupled actor--critic
method. In a standard policy gradient approach, with a single policy $\pi$
with parameters $\theta$, we compute gradient steps
of the form \citep{Williams92Reinforce}:
\begin{equation}
  \label{eq:vanilla-pg}
  \nabla_\theta J(\pi) = \sum_i \big(\nabla_{\theta} \log
  \pi(a_i|s_i)\big)\big(q_i - c(s_i)\big),
\end{equation}
where the baseline or ``critic'' $c$ can be chosen independently of the future
without introducing bias into the gradient. Recalling our previous definition of
$q_i$ as the empirical return starting from $s_i$, this form of the gradient
corresponds to a generalized advantage estimator \citep{Schulman15GAE} with
$\lambda = 1$.  Here $c$ achieves close to the optimal variance
\citep{Greensmith04PG} when it is set exactly equal to the state-value function
$V_\pi(s_i) = \expect_\pi q_i$ for the target policy $\pi$ starting in state
$s_i$.

The situation becomes slightly more complicated when generalizing to modular
policies built by sequencing subpolicies. In this case, we will have one
subpolicy per symbol but one critic per \emph{task}. This is because subpolicies
$\pi_b$ might participate in a number of composed policies $\Pi_\task$, each
associated with its own reward function $R_\task$. Thus individual subpolicies
are not uniquely identified with value functions, and the aforementioned
subpolicy-specific state-value estimator is no longer well-defined. 
We extend the actor--critic method to incorporate the decoupling of policies
from value functions by allowing the critic to vary per-sample (that is,
per-task-and-timestep) depending on the reward function with which the sample is
associated. Noting that
$\nabla_{\theta_b} J(\mathbf{\Pi}) = 
\sum_{t: b \in K_\task} \nabla_{\theta_b} J(\Pi_\task)$, i.e.\ the sum of
gradients of expected rewards across all tasks in
which $\pi_b$ participates, we have:
\begin{align}
  \label{eq:decoupled-pg}
  \nabla_\theta &J(\mathbf{\Pi}) = \sum_\task \nabla_\theta J(\Pi_\task) \nonumber \\
  &=
  \sum_\task \sum_i \big(\nabla_{\theta_b} \log
  \pi_b(a_{\task i}|s_{\task i})\big)\big(q_i - c_\task(s_{\task i})\big),
\end{align}
where each state-action pair $(s_{\task i}, a_{\task i})$ was selected by the
subpolicy $\pi_b$ in the context of the task $\task$. 

Now minimization of the gradient variance requires that each $c_\task$ actually
depend on the task identity. (This follows immediately by applying the
corresponding argument in \citet{Greensmith04PG} individually to each term in the
sum over $\task$ in \autoref{eq:decoupled-pg}.) Because the value function is
itself unknown, an approximation must be estimated from data.  Here we allow
these $c_\task$ to be implemented with an arbitrary function approximator
with parameters $\eta_\task$.
This is trained to minimize a squared
error criterion, with gradients given by
\begin{align}
  \nabla_{\eta_\task} \bigg[ -\frac{1}{2} \sum_i (&q_i - c_\task(s_i))^2 \bigg] \nonumber \\
  &= \sum_i \big( \nabla_{\eta_\task} c_\task(s_i) \big)\big(q_i - c_\task(s_i)\big).
\end{align}
Alternative forms of the advantage estimator (e.g.\ the TD residual $R_\task(s_i) +
\gamma V_\task(s_{i+1}) - V_\task(s_i)$ or any other member of the generalized
advantage estimator family)
can be easily substituted by simply maintaining one such estimator per task.
Experiments (\autoref{ssec:ablations}) show that conditioning on both the state
and the task identity results in noticeable performance improvements, suggesting
that the variance reduction provided by this objective is important for
efficient joint learning of modular policies.

The complete procedure for computing a \emph{single} gradient step is given in
\autoref{alg:inner-loop}. (The outer training loop over these steps, which is
driven by a curriculum learning procedure, is 
specified in \autoref{alg:outer-loop}.) This is an on-policy algorithm. In
each step, the agent samples tasks from a task distribution provided
by a curriculum (described in the following subsection).  The current family of
policies $\mathbf{\Pi}$ is used to perform rollouts in each sampled task,
accumulating the resulting tuples of (states, low-level actions, high-level
symbols, rewards, and task identities) into a dataset $\data$. Once $\data$
reaches a maximum size $D$, it is used to compute gradients w.r.t.\ both
policy and critic parameters, and the parameter vectors are updated accordingly.
The step sizes $\alpha$ and $\beta$ in \autoref{alg:inner-loop} can be chosen
adaptively using any first-order method. 

\subsection{Curriculum Learning}
\label{sec:curriculum}

For complex tasks, like the one depicted in \autoref{fig:tasks}b, it is
difficult for the agent to discover any states with positive reward until many
subpolicy behaviors have already been learned. It is thus a better use of the
learner's time to focus on ``easy'' tasks, where many rollouts will result
in high reward from which appropriate subpolicy behavior can be inferred. But
there is a fundamental tradeoff involved here: if the learner spends too much
time on easy tasks before being made aware of the existence of harder ones, it
may overfit and learn subpolicies that no longer generalize or exhibit the
desired structural properties.

To avoid both of these problems, we use a curriculum learning scheme
\citep{Bengio09Curriculum} that allows
the model to smoothly scale up from easy tasks to more difficult ones while
avoiding overfitting.  Initially the model is presented with tasks associated
with short sketches. Once average reward on all these tasks reaches a certain
threshold, the length limit is incremented. We assume that rewards across tasks
are normalized with maximum achievable reward $0 < q_i < 1$. Let
$\hat{\expect}r_\tau$ denote the empirical estimate of the expected reward for
the current policy on task $\task$. Then at each timestep, tasks are sampled in
proportion to $1 - \hat{\expect}r_\tau$, which by assumption must be positive.

Intuitively, the tasks that provide the strongest learning signal are those in
which (1) the agent does not on average achieve reward close to the upper bound,
but (2) many episodes result in high reward. The expected reward
component of the curriculum addresses condition (1) by ensuring that time is not
spent on nearly solved tasks, while the length bound component of the curriculum
addresses condition (2) by ensuring that tasks are not attempted until
high-reward episodes are likely to be encountered.
Experiments show that both components of this curriculum learning scheme improve
the rate at which the model converges to a good policy
(\autoref{ssec:ablations}).

The complete curriculum-based training procedure is specified in
\autoref{alg:outer-loop}. Initially, the maximum sketch length
$\ell_\textrm{max}$ is set to 1, and the curriculum initialized to sample
length-1 tasks uniformly. (Neither of the environments we consider in this paper
feature any length-1 tasks; in this case, observe that \autoref{alg:outer-loop}
will simply advance to length-2 tasks without any parameter updates.) For each
setting of $\ell_\textrm{max}$, the algorithm uses the current collection of
task policies $\mathbf{\Pi}$ to compute and apply the gradient step described in
\autoref{alg:inner-loop}. The rollouts obtained from the call to
$\textsc{train-step}$ can also be used to compute reward estimates $\hat\expect
r_\task$; these estimates determine a new task distribution for
the curriculum. The inner loop is repeated until the reward threshold
$r_\textrm{good}$ is exceeded, at which point $\ell_\textrm{max}$ is incremented
and the process repeated over a (now-expanded) collection of tasks.

\section{Experiments}

\label{sec:experiments}

We evaluate the performance of our approach
in three environments: a crafting environment, a maze navigation environment,
and a cliff traversal environment. These environments involve various kinds of
challenging low-level control: agents must learn to avoid
obstacles, interact with various kinds of objects, and relate fine-grained joint
activation to high-level locomotion goals. They also
feature hierarchical structure: most rewards are provided only after the agent 
has completed two to five high-level actions in the appropriate sequence, without any intermediate goals to indicate progress towards completion.

\subsection{Implementation}

In all our experiments, we implement each subpolicy as a feedforward neural network
with ReLU nonlinearities and a hidden layer with 128 hidden units,
and each critic as a linear
function of the current state. Each subpolicy network receives as input a set of
features describing the current state of the environment, and outputs a
distribution over actions. The agent acts at every timestep by sampling
from this distribution. 
The gradient steps given in lines 8 and 9 of \autoref{alg:inner-loop} are
implemented using \textsc{RMSProp} \citep{Tieleman12RMSProp} with a step size of 0.001 and gradient
clipping to a unit norm. We take the batch size $D$ in
\autoref{alg:inner-loop} to be 2000, and set $\gamma=0.9$ in both environments.
For curriculum learning, the improvement threshold $r_\textrm{good}$ is 
0.8.

\begin{figure}
  \centering
  \raisebox{4.5em}{(a)} \hspace{1em} \includegraphics[width=2.2in, trim=0.1in
  4.2in 4.1in 0.2in, clip]{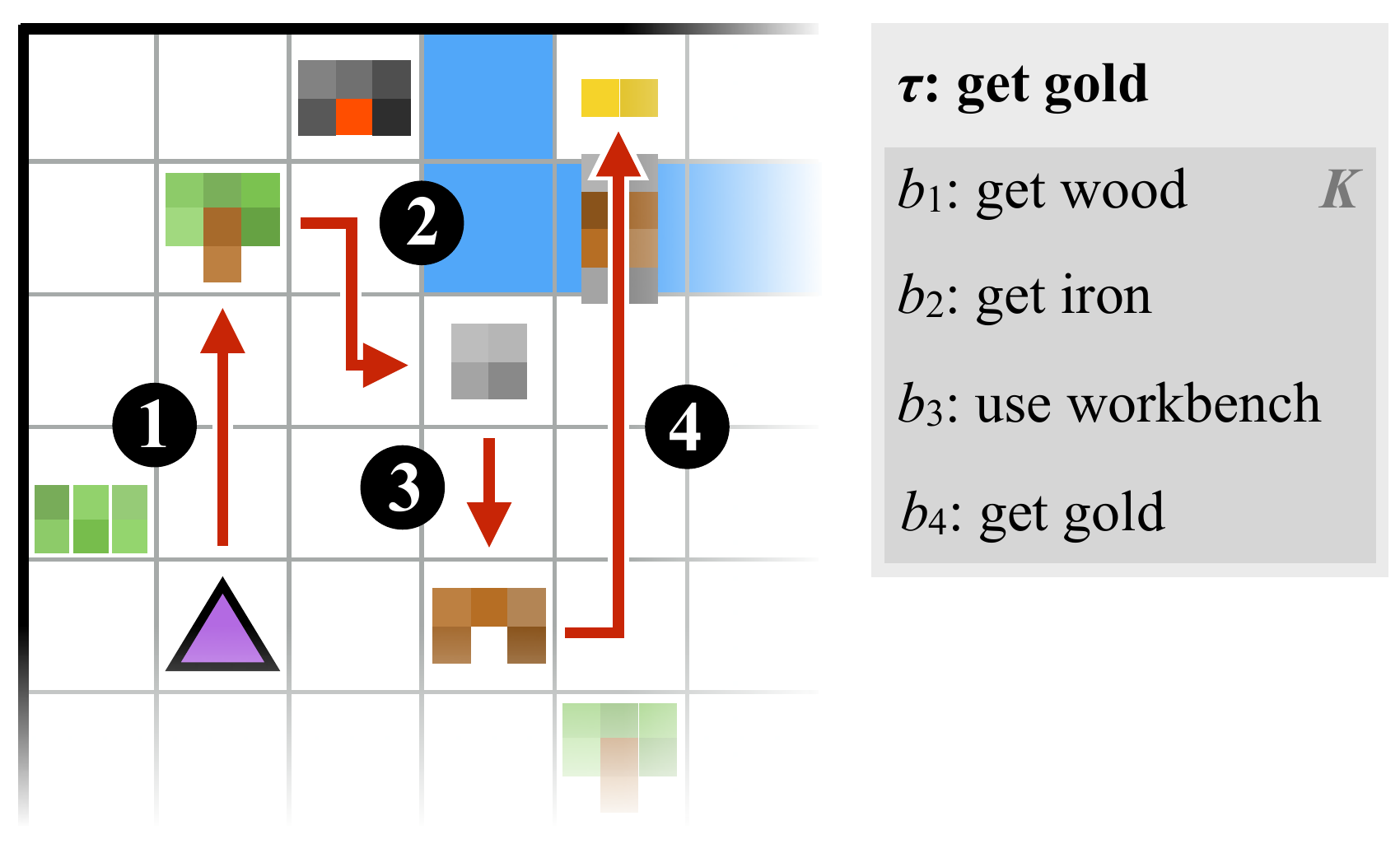} \\[1em]
  \raisebox{4.5em}{(b)} \hspace{1em} \includegraphics[width=2.2in, trim=0.1in 4.2in 4.1in 0.2in, clip]{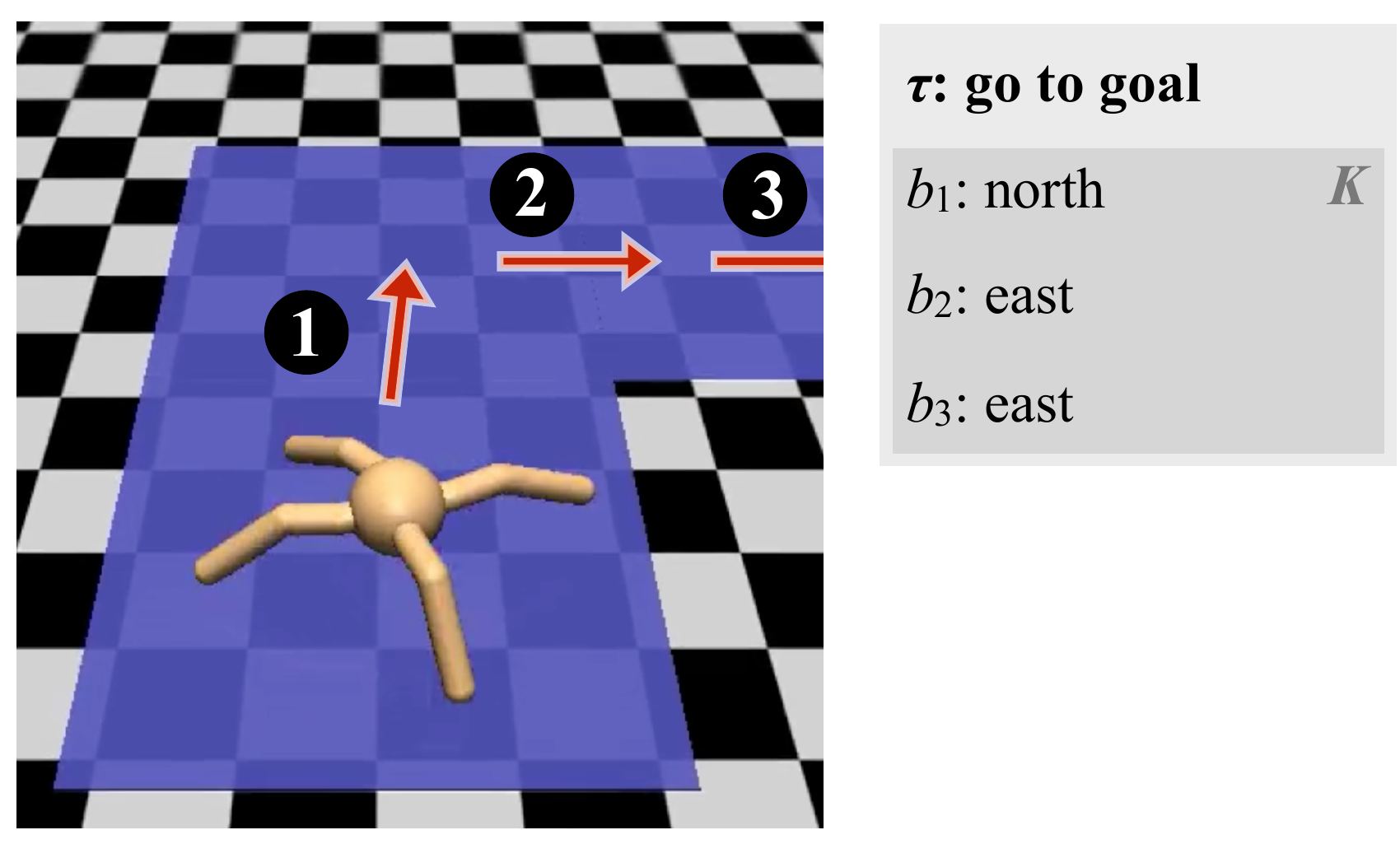}
  \caption{
      Examples from the crafting and cliff environments used in this paper. An
    additional maze environment is also investigated.
                    (a) 
    In the crafting environment, an agent seeking to pick up the gold nugget in
    the top corner must first collect wood (1) and iron (2), use a workbench to
    turn them into a bridge (3), and use the bridge to cross the water (4). 
    (b) 
    In the cliff environment, the agent must reach a goal position by traversing
    a winding sequence of tiles without falling off. Control takes place at the level
    of individual joint angles; high-level behaviors like ``move north'' must be learned.
      }
  \label{fig:tasks}
  \vspace{-1em}
\end{figure}

\subsection{Environments}
\label{ssec:envs}

\begin{figure*}
  {
  \centering
  \includegraphics[width=0.32\textwidth]{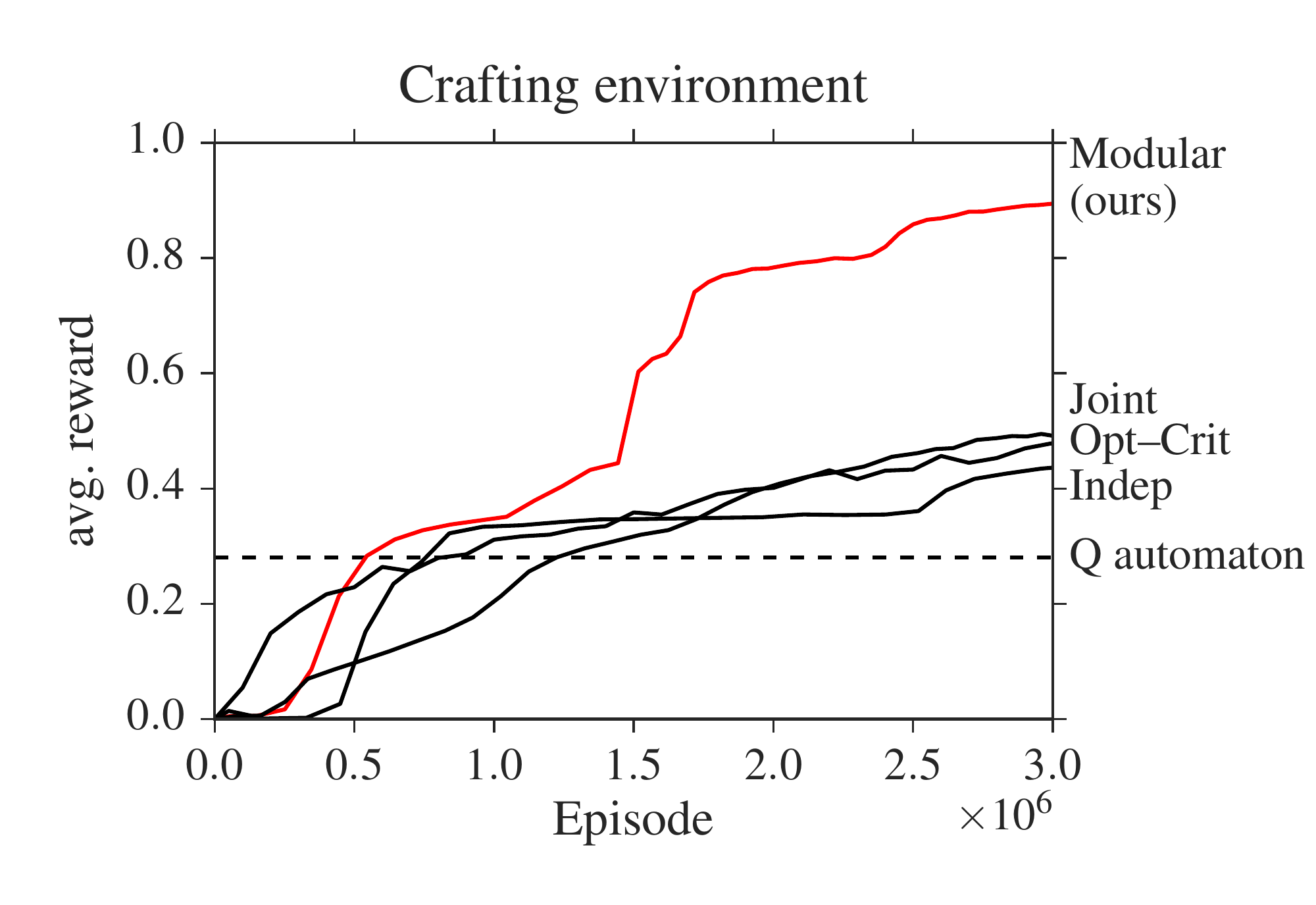}
  \includegraphics[width=0.32\textwidth]{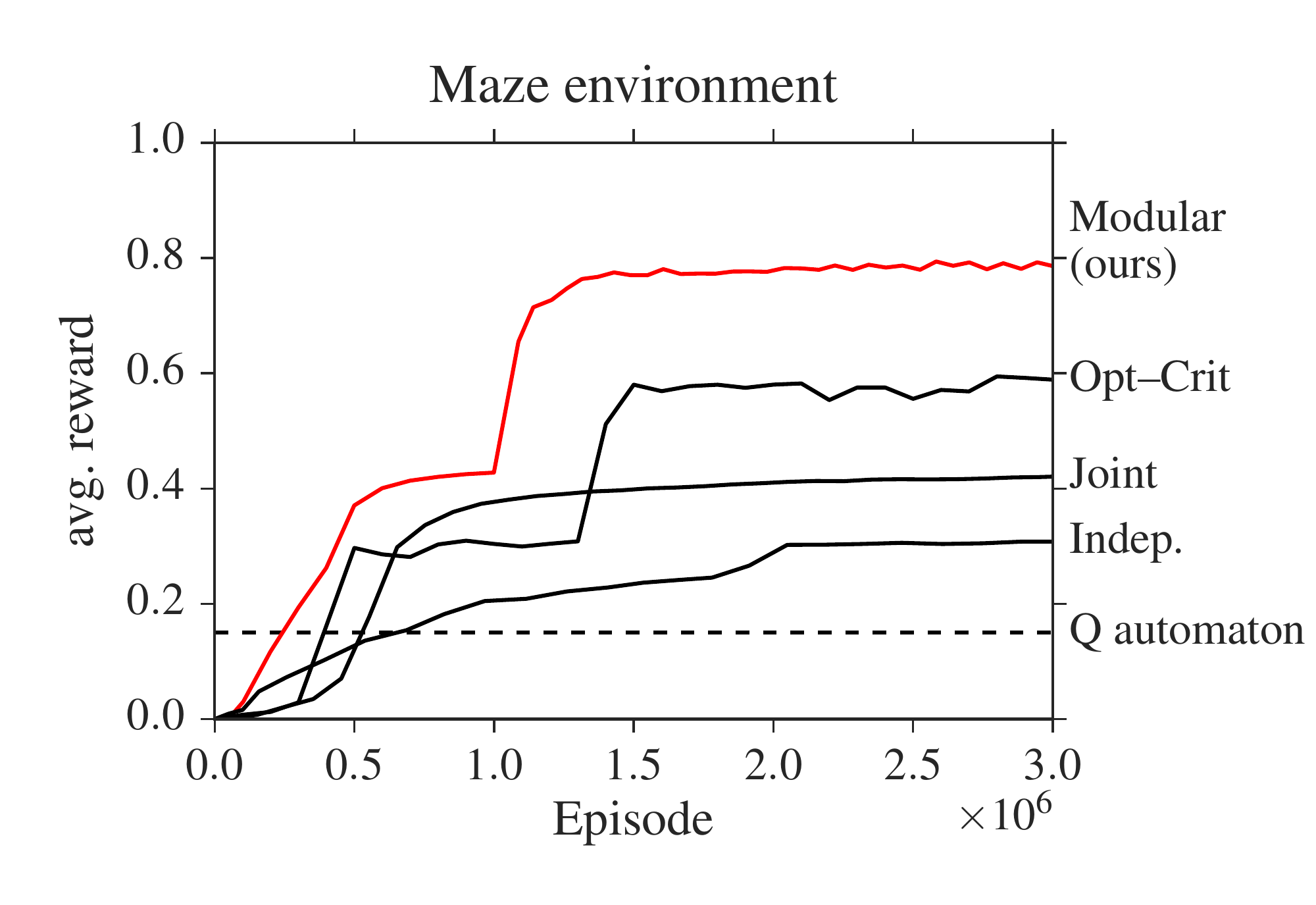}
  \includegraphics[width=0.32\textwidth]{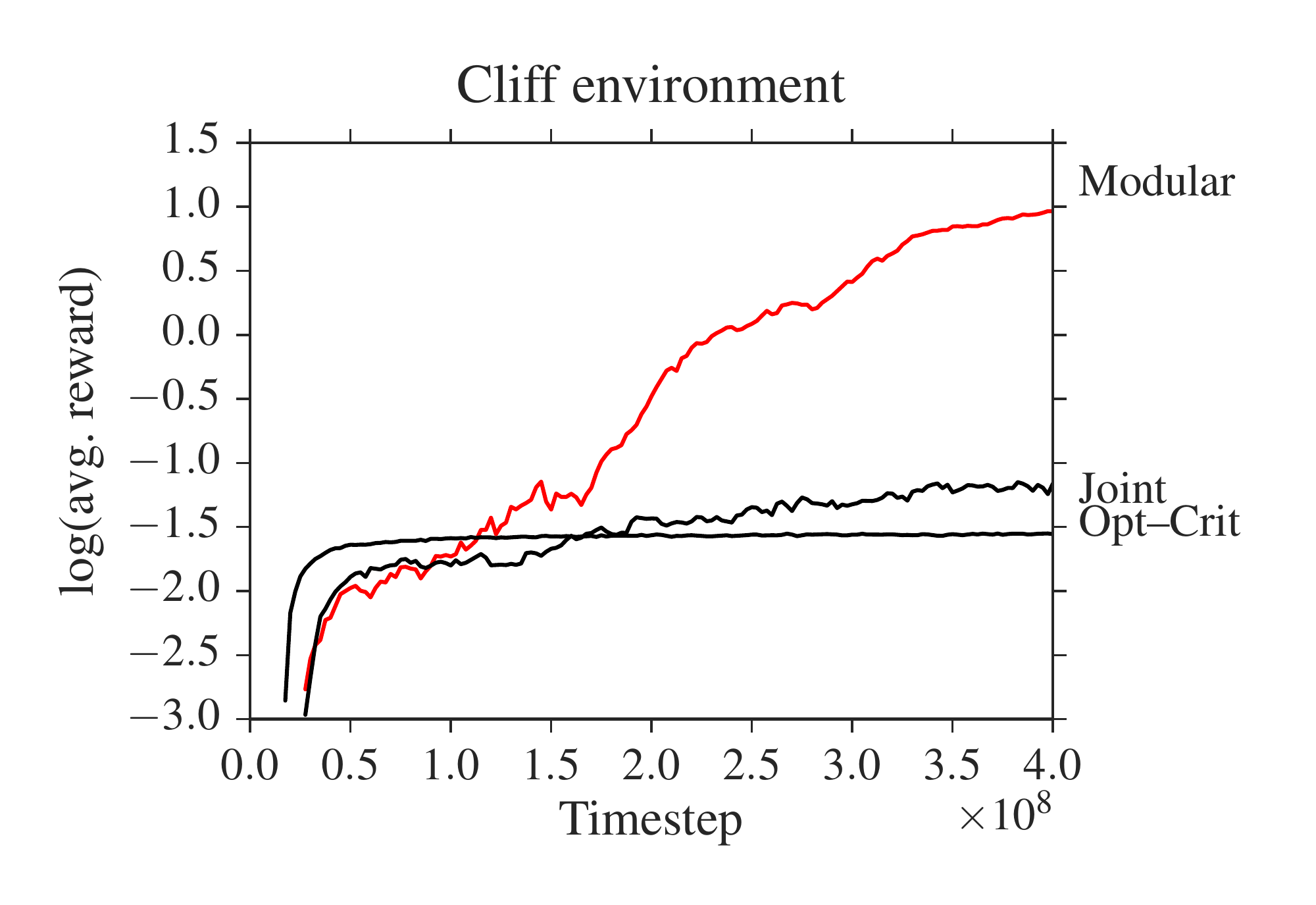} \\
  }
  \hspace{7.4em} (a) \hspace{14.3em} (b) \hspace{14.4em} (c)
  \caption{
    Comparing modular learning from sketches with standard RL baselines.
    \textbf{Modular} is the approach described in this paper, while
    \textbf{Independent} learns a separate policy for each task, \textbf{Joint}
    learns a shared policy that conditions on the task identity, \textbf{Q
    automaton} learns a single network to map from states and action symbols to Q
    values, and \textbf{Opt--Crit} is an unsupervised option learner. 
    Performance for the best iteration of the (off-policy) Q automaton is plotted.
    Performance is shown in (a) the crafting environment, (b) the maze environment,
    and (c) the cliff environment.
                            The modular approach is
    eventually able to achieve high reward on all tasks, while the baseline
    models perform considerably worse on average.
  }
  \label{fig:multitask}
  \vspace{-.5em}
\end{figure*}

\textbf{The crafting environment} (\autoref{fig:tasks}a) is
inspired by the popular game Minecraft, but is implemented in a discrete \mbox{2-D} world.
The agent may interact with objects in the world by facing them and
executing a special \textsc{use} action. Interacting with raw materials
initially scattered around the environment causes them to be added to an
inventory. Interacting with different crafting stations causes objects in the
agent's inventory to be combined or transformed. Each task in
this game corresponds to some crafted object the agent must produce; the most
complicated goals require the agent to also craft intermediate ingredients, and
in some cases build tools (like a pickaxe and a bridge) to reach ingredients
located in initially inaccessible regions of the environment.

\textbf{The maze environment} (not pictured)
corresponds closely to the the ``light world'' described by
\citet{Konidaris07Skills}. The agent is placed in a discrete world consisting of
a series of rooms, some of which are connected by doors. Some doors require that
the agent first pick up a key to open them. For our experiments, each task
corresponds to a goal room (always at the same position relative to the agent's
starting position) that the agent must reach by navigating through a sequence of
intermediate rooms. The agent has one sensor on each side of its body, which
reports the distance to keys, closed doors, and open doors in the corresponding
direction. Sketches specify a particular sequence of directions for the agent
to traverse between rooms to reach the goal. 
The sketch always corresponds to a viable traversal from the
start to the goal position, but other (possibly shorter) traversals may also exist.

\textbf{The cliff environment} (\autoref{fig:tasks}b) is
intended to demonstrate the applicability of our approach
to problems involving high-dimensional continuous control. In this
environment, a quadrupedal robot \cite{Schulman15TRPO}
is placed on a variable-length winding
path, and must navigate to the end without falling off.
This task is designed to provide a substantially more challenging
RL problem, due to the fact that the walker must learn the low-level
walking skill before it can make any progress, but has simpler
hierarchical structure than the crafting environment. The
agent receives a small reward for making progress toward the
goal, and a large positive reward for reaching the goal square,
with a negative reward for falling off the path.

A 
listing of tasks and sketches is given in \autoref{app:tasks}.

\subsection{Multitask Learning}
\label{ssec:multitask}

The primary experimental question in this paper is whether the extra structure
provided by policy sketches alone
is enough to enable fast learning of coupled
policies across tasks. 
We aim to explore the differences between the approach described in
\autoref{sec:learning} and relevant prior work that performs
either unsupervised or weakly supervised multitask learning of hierarchical policy structure. Specifically, we compare our \textbf{modular} to approach to: \\[-1.5em]
\begin{enumerate}
	\item Structured hierarchical reinforcement learners:
    	\begin{enumerate}
          \item[(a)] the fully unsupervised \textbf{option--critic} algorithm of \citet{Bacon15OptionCritic}
          \item[(b)]  a \textbf{Q automaton} that attempts to explicitly represent the Q function for each task / subtask combination (essentially a HAM \citep{Andre02ALISPAbstraction} with a deep state abstraction function)
        \end{enumerate}
	\item Alternative ways of incorporating sketch data into standard policy 
    	gradient methods:
    	\begin{enumerate}
          \item[(c)] learning an \textbf{independent} policy for each task
          \item[(d)] learning a \textbf{joint} policy across all tasks, conditioning
          directly on both environment features and a representation of the complete
          sketch\\[-1.5em]
        \end{enumerate}
\end{enumerate}

The joint and independent models performed best when trained
with the same curriculum described in \autoref{sec:curriculum}, while the
option--critic model performed best with a length--weighted curriculum that
has access to all tasks from the beginning of training.

Learning curves for baselines and the modular model are shown in
\autoref{fig:multitask}. It can be seen that in all environments, our approach
substantially outperforms the baselines: it induces policies with substantially
higher average reward and converges more quickly than the policy gradient
baselines. It can further be seen in \autoref{fig:multitask}c that after
policies have been learned on simple tasks, the model is able to rapidly adapt
to more complex ones, even when the longer tasks involve high-level actions not
required for any of the short tasks
(\autoref{app:tasks}).

Having demonstrated the overall effectiveness of our approach, our remaining
experiments explore (1) the importance of various components of the training
procedure, and (2) the learned models' ability to generalize or adapt to
held-out tasks.  For compactness, we restrict our consideration on the crafting
domain, which features a larger and more diverse range of tasks and high-level
actions.

\begin{figure}[t]
  \vspace{-.5em}
  \strut
  \footnotesize
  \hspace{-10pt}
  \includegraphics[height=3.3cm,trim=25pt 0pt 10pt 0pt,clip]{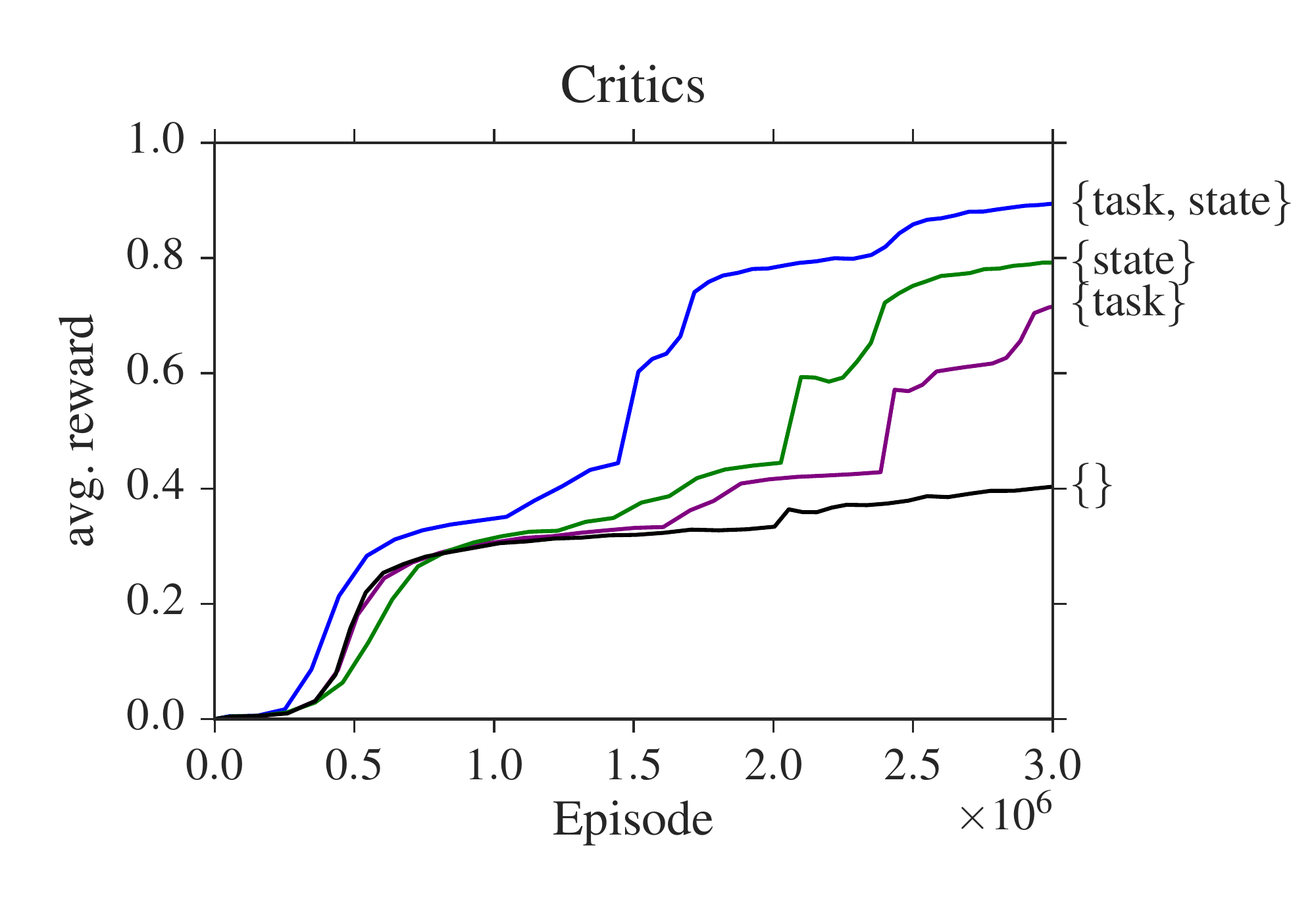}~~~~  \includegraphics[height=3.3cm,trim=50pt 0pt 25pt 0pt,clip]{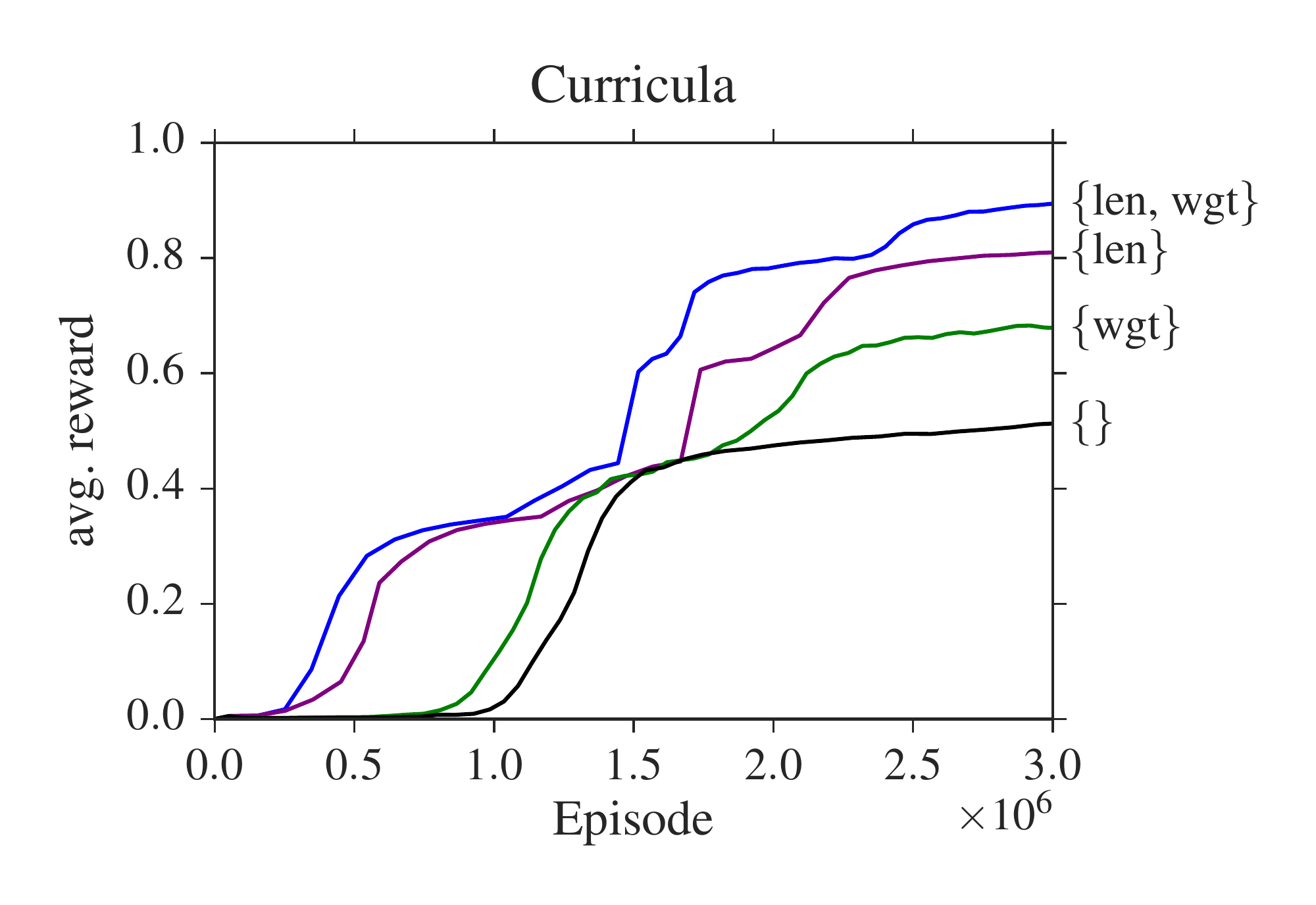}
  \\[-0.5em]\strut\hspace{5.3em}(a)\hspace{13em}(b)
  \\[-1.5em]

  \begin{center}
    \includegraphics[height=3.3cm,trim=0pt 0pt 1cm 0pt,clip]{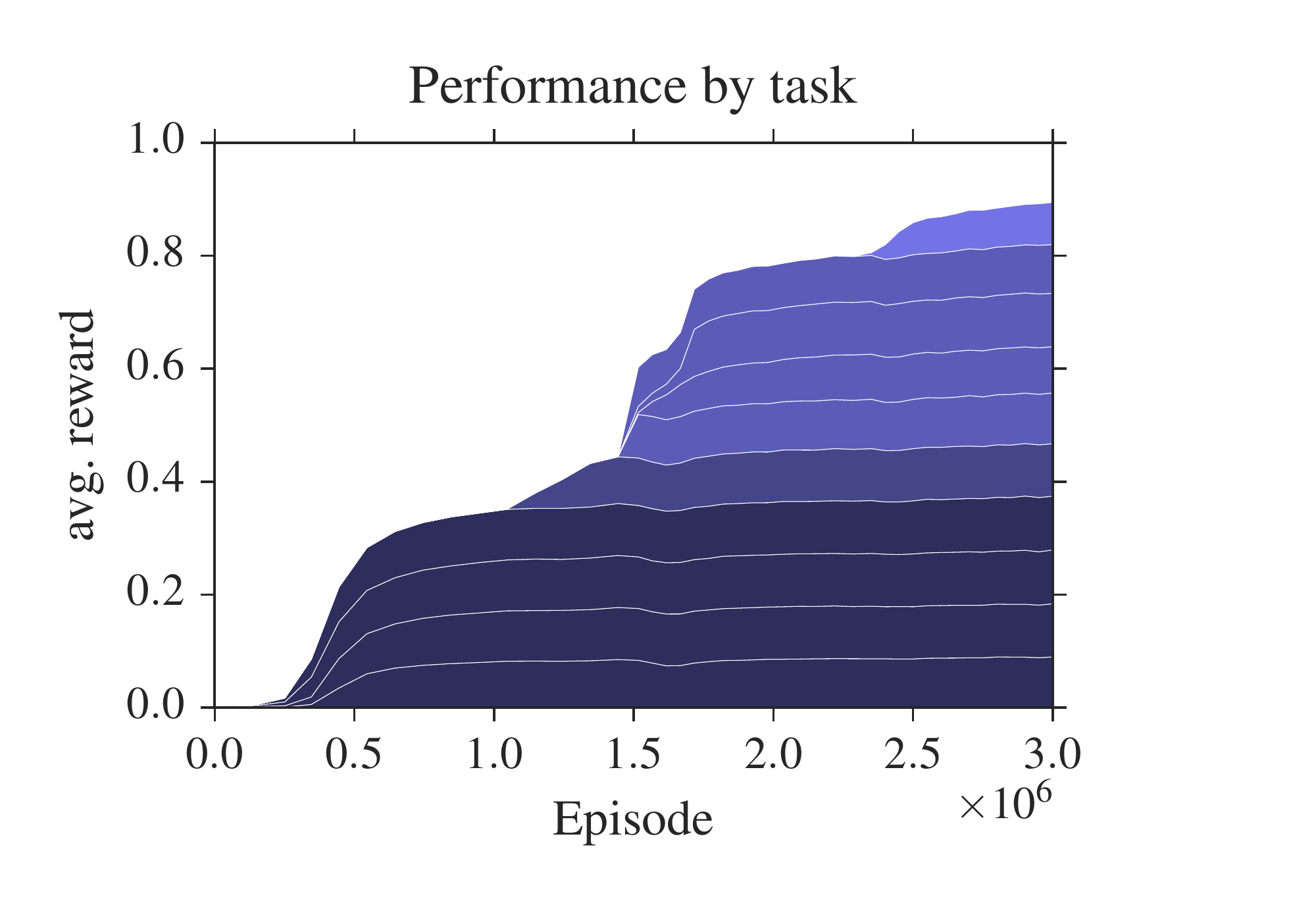}
    \\[-0.5em]
    (c)
  \end{center}
  \vspace{-1em}

  \caption{
    Training details in the crafting domain. (a) Critics: lines labeled ``task'' include a
    baseline that varies with task identity, while lines labeled ``state''
    include a baseline that varies with state identity.  Estimating a
    baseline that depends on both the representation of the current state and
    the identity of the current task is better than either alone or a constant
    baseline.  (b) Curricula: lines labeled ``len'' use a curriculum
    with iteratively increasing sketch lengths, while lines labeled ``wgt'' sample
    tasks in inverse proportion to their current reward. Adjusting the sampling
    distribution based on both task length and performance return improves convergence.
    (c) Individual task performance. Colors correspond to task length.
    Sharp steps in the learning curve correspond to
    increases of $\ell_\textrm{max}$ in the curriculum. 
  }
  \label{fig:ablations}
  \vspace{-1em}
\end{figure}

\subsection{Ablations}
\label{ssec:ablations}

In addition to the overall modular parameter-tying structure induced by our
sketches, the key components of our training procedure are the decoupled critic
and the curriculum. Our next experiments investigate the extent to which these
are necessary for good performance.

To evaluate the the critic, we consider three ablations: (1) removing the
dependence of the model on the environment state, in which case the baseline is
a single scalar per task; (2) removing the  dependence of the model on the task,
in which case the baseline is a conventional generalized advantage estimator;
and (3) removing both, in which case the baseline is a single scalar, as in a
vanilla policy gradient approach. Results are shown in \autoref{fig:ablations}a.
Introducing both state and task dependence into the baseline leads to faster
convergence of the model: the approach with a constant baseline achieves less
than half the overall performance of the full critic after 3 million episodes.
Introducing task and state dependence independently improve this performance;
combining them gives the best result.

We also investigate two aspects of our curriculum learning scheme: starting with
short examples and moving to long ones, and sampling tasks in inverse proportion
to their accumulated reward. Experiments are shown in \autoref{fig:ablations}b.
\added{
Both components help;
prioritization by both length and weight gives the best 
results.
}

\subsection{Zero-shot and Adaptation Learning}
\label{ssec:generalization}

In our final experiments, we consider the model's ability to generalize 
beyond the standard training condition.
We first consider two tests of generalization: a
\textbf{zero-shot} setting, in which the model is provided a sketch for the new
task and must immediately achieve good performance, and a \textbf{adaptation}
setting, in which no sketch is provided and the model must learn the form of a
suitable sketch via interaction in the new task.

We hold out two length-four tasks from the full inventory used in
\autoref{ssec:multitask}, and train on the remaining tasks. For zero-shot
experiments, we simply form the concatenated policy described by the sketches of
the held-out tasks, and repeatedly execute this policy (without learning) in
order to obtain an estimate of its effectiveness.  For adaptation experiments,
we consider ordinary RL over high-level actions $\options$ rather than
low-level actions $\actions$, implementing the high-level learner with the same agent architecture
as described in \autoref{ssec:model}. Note that the Independent and Option--Critic models cannot
be applied to the zero-shot evaluation, while the Joint model cannot be
applied to the adaptation baseline (because it depends on pre-specified sketch
features).  Results are shown in \autoref{tab:generalization}. The held-out
tasks are sufficiently challenging that the baselines are unable to obtain more
than negligible reward: in particular,  the joint model overfits to the training 
tasks and cannot generalize to new  sketches, while the independent model cannot 
discover enough of a reward signal to learn in the adaptation setting.
The modular model does comparatively well: individual subpolicies succeed
in novel zero-shot configurations (suggesting that they have in fact discovered
the behavior suggested by the semantics of the sketch) and provide a suitable
basis for adaptive discovery of new high-level policies.

\begin{table}[t]
  \centering
  {\footnotesize
	\begin{tabular}{lccc}
    \toprule
    Model & Multitask & 0-shot & Adaptation \\
    \midrule
    Joint & .49 & .01 & -- \\
    Independent & .44 & -- & .01 \\
    Option--Critic & .47 & -- & .42 \\
    Modular (ours) & \bf .89 & \bf .77 & \bf .76 \\
    \bottomrule
  \end{tabular}
  }
  \caption{
    Accuracy and generalization of learned models in the crafting domain. The table
    shows the task completion rate for each approach after convergence under
    various training conditions.
    \textbf{Multitask} is the multitask training condition described in 
    \autoref{ssec:multitask}, while \textbf{0-Shot} and \textbf{Adaptation} are the 
    generalization experiments described in \autoref{ssec:generalization}.
    Our modular approach consistently achieves the best performance.
  }
  \label{tab:generalization}
  \vspace{-1em}
\end{table}

\section{Conclusions}

We have described an approach for multitask learning of deep multitask policies
guided by symbolic policy sketches. By associating each symbol appearing in a
sketch with a modular neural subpolicy, we have shown that it is possible to
build agents that share behavior across tasks in order to achieve success in
tasks with sparse and delayed rewards. This process induces an inventory of
reusable and interpretable subpolicies which can be employed for zero-shot
generalization when further sketches are available, and hierarchical
reinforcement learning when they are not. Our work suggests that these sketches,
which are easy to produce and require no grounding in the environment, provide
an effective scaffold for learning hierarchical policies from minimal
supervision.  

\section*{Acknowledgments}
JA is supported by a Facebook Graduate Fellowship and a Berkeley AI / Huawei
Fellowship.

\bibliography{jacob}
\bibliographystyle{icml2017}

\twocolumn[
\appendix

\section{Tasks and Sketches}
\label{app:tasks}

The complete list of tasks, sketches, and symbols is given below. Tasks marked
with an asterisk$^*$ are held out for the generalization experiments described
in \autoref{ssec:generalization}, but included in the multitask training
experiments in Sections \ref{ssec:multitask} and \ref{ssec:ablations}. \\

{\ttfamily\footnotesize
\begin{tabular}{l|lllll}
  \toprule
  \rmfamily Goal & \rmfamily Sketch \\
  \midrule
  \multicolumn{4}{l}{\rmfamily \bfseries Crafting environment} \\
  \midrule
  make plank   & get wood  & use toolshed \\
  make stick   & get wood  & use workbench \\
  make cloth   & get grass & use factory \\
  make rope    & get grass & use toolshed \\
  make bridge  & get iron  & get wood      & use factory \\
  make bed$^*$ & get wood  & use toolshed  & get grass & use workbench \\
  make axe$^*$ & get wood  & use workbench & get iron  & use toolshed \\
  make shears  & get wood  & use workbench & get iron  & use workbench \\
  get gold     & get iron  & get wood      & use factory   & use bridge \\
  get gem      & get wood  & use workbench & get iron  & use toolshed & use axe \\
  \midrule
  \multicolumn{4}{l}{\rmfamily \bfseries Maze environment} \\
  \midrule
  room 1  & left & left \\
  room 2  & left & down \\
  room 3  & right & down\\
  room 4  & up & left \\
  room 5  & up & right \\
  room 6  & up & right & up \\
  room 7  & down & right & up \\
  room 8  & left & left & down \\
  room 9  & right & down & down \\
  room 10  & left & up & right \\
  \midrule
  \multicolumn{4}{l}{\rmfamily \bfseries Cliff environment} \\
  \midrule
  path 0 & north \\
  path 1 & east \\
  path 2 & south \\
  path 3 & west \\
  path 4 & west & south \\
  path 5 & west & north & north \\
  path 6 & north & east & north \\
  path 7 & west & north \\
  path 8 & east & south \\
  path 9 & north & west & west \\
  path 10 & east & north & east \\
  path 11 & south & east \\
  path 12 & south & west \\
  path 13 & south & south \\
  path 14 & south & south & west \\
  path 15 & east & south & south \\
  path 16 & east & east \\
  path 17 & east & north \\
  path 18 & north & east \\
  path 19 & west & west \\
  path 20  & north & north \\
  path 21  & north & west \\
  path 22  & west & west & south \\
  path 23  & south & east & south \\
  \bottomrule
\end{tabular}
}

]

\end{document}